\documentclass[letterpaper]{article} % DO NOT CHANGE THIS
\usepackage{aaai24}
\usepackage{times}  % DO NOT CHANGE THIS
\usepackage{helvet}  % DO NOT CHANGE THIS
\usepackage{courier}  % DO NOT CHANGE THIS
\usepackage[hyphens]{url}  % DO NOT CHANGE THIS
\usepackage{graphicx} % DO NOT CHANGE THIS
\usepackage{natbib}  % DO NOT CHANGE THIS AND DO NOT ADD ANY OPTIONS TO IT
\usepackage{caption} % DO NOT CHANGE THIS AND DO NOT ADD ANY OPTIONS TO IT
\usepackage{algorithm}
\usepackage{algorithmic}
\usepackage{newfloat}
\usepackage{listings}
\DeclareCaptionStyle{ruled}{labelfont=normalfont,labelsep=colon,strut=off} % DO NOT CHANGE THIS
\floatstyle{ruled}
\newfloat{listing}{tb}{lst}{}
\floatname{listing}{Listing}
\usepackage{xspace}
\makeatletter
\DeclareRobustCommand\onedot{\futurelet\@let@token\@onedot}
\def\@onedot{\ifx\@let@token.\else.\null\fi\xspace}
\def\eg{\emph{e.g}\onedot} 
\def\ie{\emph{i.e}\onedot} 
 
\def\etc{\emph{etc}\onedot} 
\def\wrt{w.r.t\onedot} 
\def\etal{\emph{et al}\onedot}
\makeatother
\usepackage[switch]{lineno}  % add line numbers
\usepackage{color}
\usepackage{leftidx}
\usepackage{amsmath}
\usepackage{amssymb}
\usepackage{mathtools}
\usepackage{bbding}
\usepackage{booktabs}
\usepackage{multirow}
\usepackage{makecell}
\usepackage{enumitem}
\usepackage{dsfont}
\usepackage{array}
\title{SiMA-Hand: Boosting 3D Hand-Mesh Reconstruction by \\Single-to-Multi-View Adaptation}
\author{
    Yinqiao Wang\textsuperscript{\rm 1,\rm 2},\hspace{0.05cm}
    Hao Xu\textsuperscript{\rm 1,\rm 2},\hspace{0.05cm}
    Pheng-Ann Heng\textsuperscript{\rm 1,\rm 2},\hspace{0.05cm}
    Chi-Wing Fu\textsuperscript{\rm 1,\rm 2}
}
\affiliations{
    \textsuperscript{\rm 1}Department of Computer Science and Engineering, CUHK\\
    \textsuperscript{2}Institute of Medical Intelligence and XR, CUHK\\
    \{yqwang, xuhao, pheng, cwfu\}@cse.cuhk.edu.hk
}

\usepackage{bibentry}
\begin{document}

\maketitle
% \linenumbers

\begin{abstract}
Estimating 3D hand mesh from RGB images is a long-standing track, in which occlusion is one of the most challenging problems.
Existing attempts towards this task often fail when the occlusion dominates the image space.
In this paper, we propose SiMA-Hand, aiming to boost the mesh reconstruction performance by \textbf{Si}ngle-to-\textbf{M}ulti-view \textbf{A}daptation.
First, we design a multi-view hand reconstructor to fuse information across multiple views by holistically adopting feature fusion at image, joint, and vertex levels.
Then, we introduce a single-view hand reconstructor equipped with SiMA. Though taking only one view as input at inference, the shape and orientation features in the single-view reconstructor can be enriched by learning non-occluded knowledge from the extra views at training, enhancing the reconstruction precision on the occluded regions.
We conduct experiments on the Dex-YCB and HanCo benchmarks with challenging object- and self-caused occlusion cases, manifesting that SiMA-Hand consistently achieves superior performance over the state of the arts.
Code will be released on~\url{https://github.com/JoyboyWang/SiMA-Hand_Pytorch}.

\end{abstract}

\section{Introduction}
Hand-mesh reconstruction from monocular RGB image is a fundamental and significant task.
It has great potential for many applications,~\eg, augmented reality, human-computer interactions,~\etc.
Yet, to obtain high-quality reconstruction, a common but very challenging scenario is that self- or object-caused occlusions often occur, making it hard to infer the shape and pose of the hand in the occluded parts.

Recently, several methods have been proposed to alleviate the ambiguity of occlusion. 
Some~\cite{chu2017multi, zhu2019robust, zhou2020occlusion} attempt to extract occlusion-robust features by adopting the spatial attention mechanism. However, their performance degrades significantly when the occluded regions dominate the image. 
Others~\cite{hasson2020leveraging, yang2020seqhand, Xu_2023_CVPR} leverage non-occluded information from multi-frame inputs. 
Yet, they often fail when the occlusion persists over nearby frames; also, cross-frame processing is typically more time-consuming.
\begin{figure}[t]
    \centering
    \includegraphics[width=0.99\linewidth]{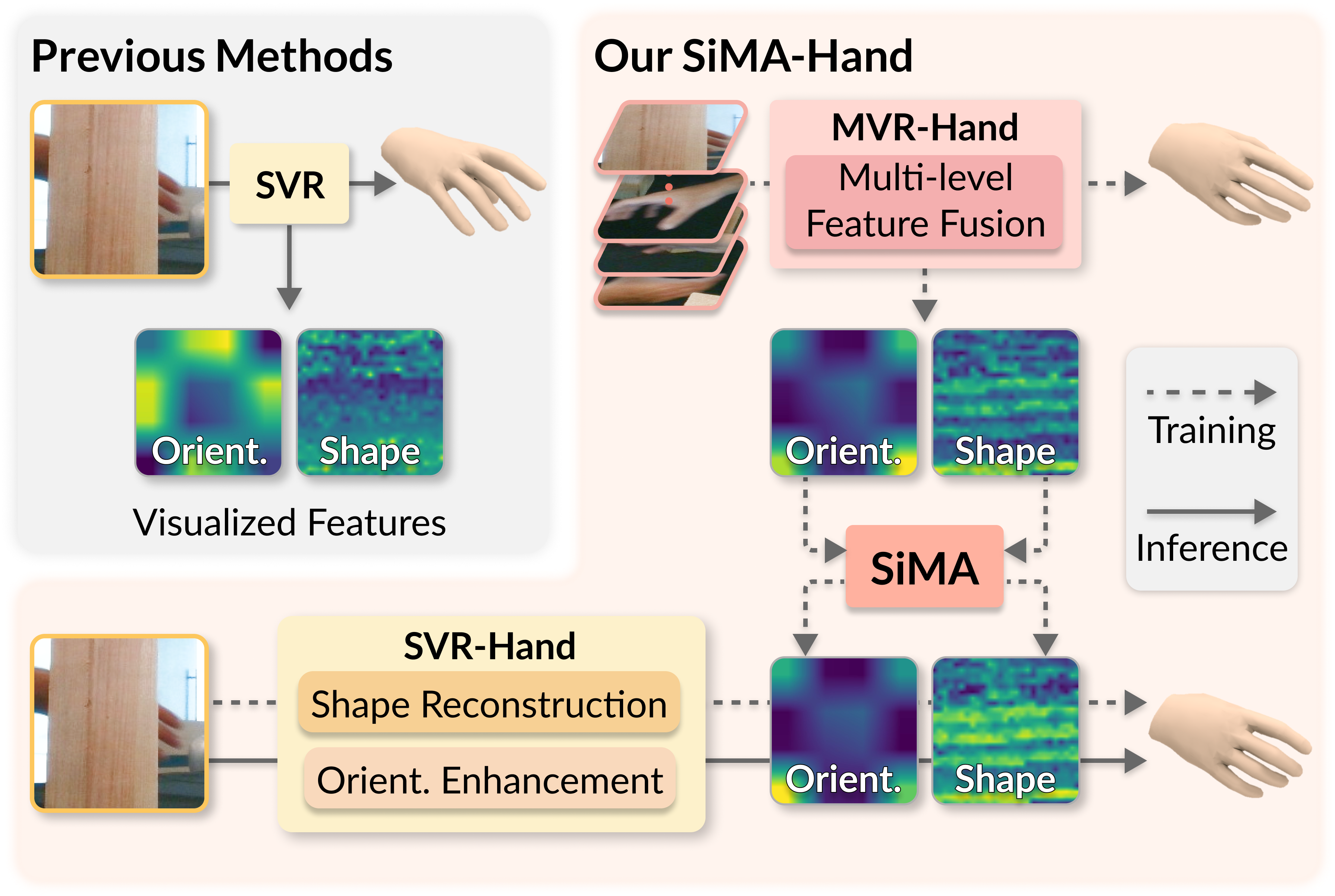}
    \caption{Framework comparison between our SiMA-Hand and previous methods. We exploit non-occluded information from the multi-view reconstructor (MVR) and carefully adopt it to the single-view reconstructor (SVR) through the proposed single-to-multi-view adaptation techniques.
    So, our SVR can learn to extract orientation and shape features, following the MVR's, and produce higher quality meshes.
    }
    \label{fig:teaser}
\end{figure}

To address this issue, another approach is to reconstruct the hand from multi-view images.
Leveraging a neural network, we can effectively build a more complete hand geometry by utilizing 3D cues from different views. 
POEM~\cite{yang2023poem} attempts to reconstruct the hand mesh based on the interaction between mesh vertices and 3D points in the intersection area of different camera frustums.
Though it helps improve the performance, the computational cost is inevitably largely increased.
Besides, the approach requires a multi-camera set up, which is tedious and expensive to prepare, compared with using commodity hand-held cameras.

In this paper, we present \textbf{SiMA-Hand}, a new framework formulated with \textbf{Si}ngle-to-\textbf{M}ulti-view \textbf{A}daptation to boost the hand-mesh reconstruction quality from single-view input.
Fig.~\ref{fig:teaser} overviews the method design. 
Our key insight is that human hands have limited feasible poses; given a single view, even the hand is partially occluded, we may still try to infer its 3D structure in some other views.
With this preliminary, we leverage the knowledge of different views from the multi-view reconstructor (MVR) during the training, to enrich the features extracted by the single-view reconstructor (SVR) from a single-view input; see again Fig.~\ref{fig:teaser}.
Using this design, during the inference, the trained SVR can perform 3D hand-mesh reconstruction with only a single view as its input, yet appearing to have a multi-view input.

Technical-wise, we first design a multi-view 3D hand reconstructor, named MVR-Hand.
Since all views of one hand possess different global orientations while sharing the same shape, to enable better information utilization from different views, we decouple the estimation of view-dependent orientation and view-independent shape into two separate branches, following~\cite{Xu_2023_CVPR}.
After that, we holistically design the multi-view fusion at different levels to exploit and fuse information across views. 
Specifically, considering that the hand orientation is related to the global picture of the entire hand, we formulate the image- and joint-level 2D feature fusion modules: 
the former enables knowledge exchange among views whereas the latter propagates joint information from the shape to the orientation branch. 
For the hand shape, as it is view-independent at the canonical pose, we design vertex-level 3D feature fusion to adaptively fuse the features of the coarse hand vertices. 

Second, we introduce our single-view reconstructor, named SVR-Hand, equipped with the single-to-multi-view adaptation techniques to leverage intermediate information of the MVR-Hand to improve the single-view reconstruction quality. 
As regressing the hand shape at the canonical pose makes the 3D shape features less related to the camera pose, directly using the multi-view 3D vertex features as guidance is highly desirable.
However, it is non-trivial to leverage the multi-view knowledge to improve the view-dependent hand-orientation precision of the target view. 
To this end, we propose the hand-orientation feature enhancement module in our SVR-Hand. 
It helps the network to effectively exploit cues from the hand-shape branch for the hand-orientation features to simulate the absent information provided by the multi-view inputs during the training.

Our main contributions are summarized below:
\begin{itemize}
    \item We introduce SiMA-Hand, a novel framework for boosting the 3D hand-mesh reconstruction with a single-to-multi-view adaptation. To our best knowledge, this is the first work that leverages multi-view information during training to improve single-view hand reconstruction.
    \item We design variant feature fusion and enhancement modules. The former is distributed at image/joint/vertex levels, aggregating useful information from multiple views to promote multi-view reconstruction, whereas the latter aims to enrich single-view orientation and shape features to perform better adaptation.
    \item We demonstrate the effectiveness of SiMA-Hand, both quantitatively and qualitatively, on two widely-used benchmarks, Dex-YCB and HanCo, showing the state-of-the-art performance of our method.
\end{itemize}

\section{Related Work}
\paragraph{Single-view hand-mesh reconstruction.}
Single-view methods can be divided into two main categories,~\ie, depth-based and RGB-based, according to the input type.

Early depth-based methods~\cite{tan2016fits, taylor2014user, khamis2015learning} reconstruct the 3D hand mesh by iteratively fitting it to the depth image. 
With their strong learning ability, deep neural networks are often adopted by recent works~\cite{malik2020handvoxnet, mueller2019real, wan2020dual} as feature extractors.

For RGB-based methods, most works formulate hand reconstruction as a problem of regressing the MANO coefficients~\cite{zhang2019end, zhou2020monocular, yang2020bihand, hasson2019learning, zhang2021hand, boukhayma20193d, baek2019pushing, zimmermann2019freihand, chen2021model, zhao2021travelnet, cao2021reconstructing, baek2020weakly, jiang2021hand, liu2021semi, zhang2021interacting, tse2022collaborative}.
Other widely-used representations include voxel-based~\cite{iqbal2018hand, moon2020i2l, moon2020interhand2, yang2021semihand}, implicit-function-based~\cite{mescheder2019occupancy}, and vertex-based~\cite{ge20193d, kulon2020weakly, chen2021camera, lin2021end, lin2021mesh}. However, they rarely consider occlusions and often fail in challenging cases.
HandOccNet~\cite{park2022handoccnet} implicitly handles occlusion for hand-object interaction scenes by employing self- and cross-attention.
Yet, hand reconstruction from single-view input becomes severely ill-posed when the occlusions dominate the image space. 

In this work, we propose to address the occlusion by leveraging useful information in multi-view inputs as guidance when training the single-view reconstructor, such that we can boost its reconstruction quality during the inference.

\paragraph{Multi-frame hand-mesh reconstruction.}
As single-view images offer limited content, some works attempt to exploit multi-frame inputs to improve the robustness and alleviate the ill-posed problem. 
SeqHAND~\cite{yang2020seqhand} utilizes Convolutional LSTM~\cite{shi2015convolutional} to improve the temporal coherence of 3D hand-pose estimation. 
\cite{chen2021temporal} adopts a self-supervised approach to predict hand meshes from a video input.
To address occlusion-caused ambiguity,~\cite{wen2023hierarchical} design a hierarchical network to utilize features of sequential frames. 
Very recently, H2ONet~\cite{Xu_2023_CVPR} exploits non-occluded information simultaneously at finger and hand levels from neighboring frames. 
Yet, the performance may degrade when a portion of the hand is consistently occluded, as the extra frames give only limited additional knowledge.
\begin{figure*}[t]
    \centering
    \includegraphics[width=.91\linewidth]{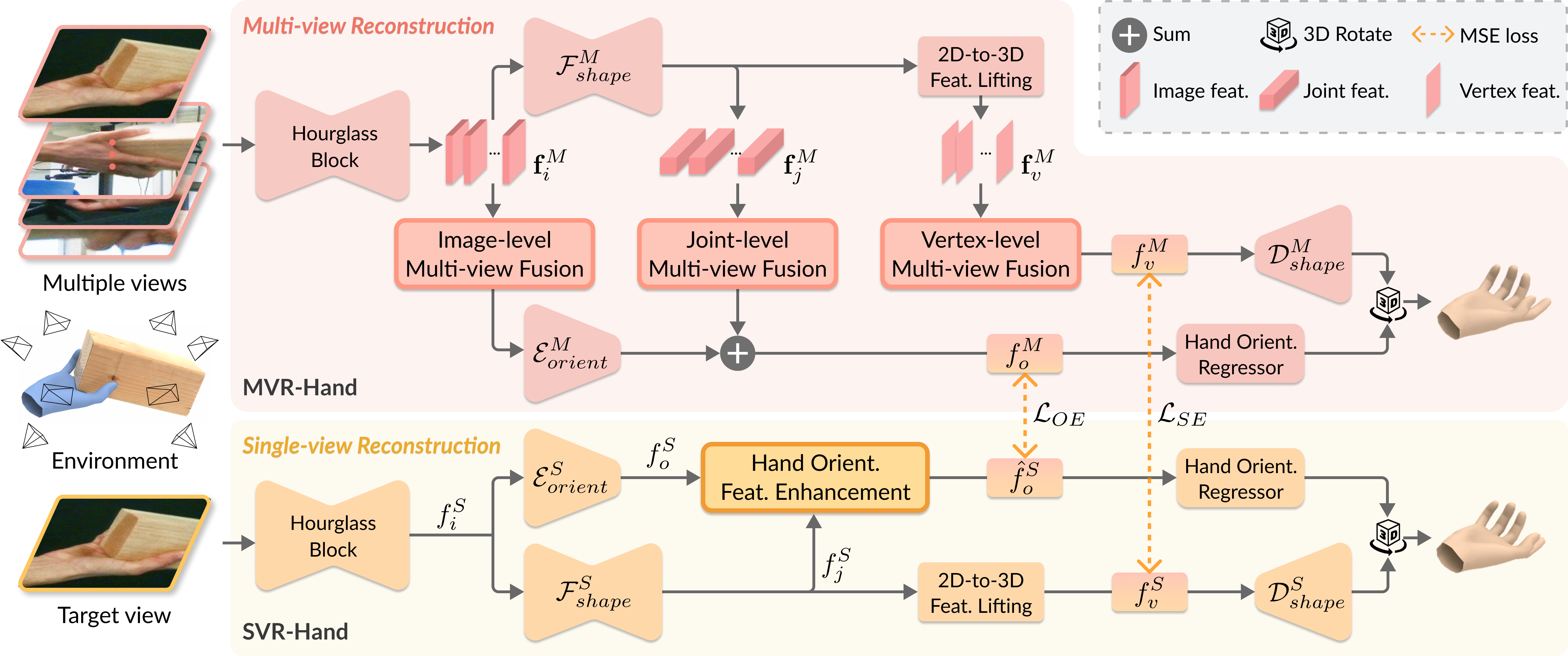}
    \caption{The architecture of SiMA-Hand: (i) the MVR-Hand takes multiple views of the hand as input for 3D hand-mesh reconstruction by fusing multi-view features at image, joint, and vertex levels; and (ii) the SVR-Hand takes only one view as input and learns to output a high-quality 3D hand mesh, even under a severely-occluded situation (see the target view at lower left), with both shape and orientation feature enhancement from the MVR-Hand.
    In the mathematical notations, SVR and MVR are denoted by superscripts $S$ and $M$, respectively.
    Also, $f$ denotes single-view or fused features, $\mathbf{f}$ denotes multi-view features, whereas subscripts $i$, $j$, $v$, and $o$ denote image, joint, vertex, and orientation, respectively.}
    \label{fig:pipeline}
\end{figure*}

\paragraph{Multi-view hand reconstruction.}
Though multi-view reconstruction for general objects is a long-standing research topic, few methods are designed for hand.
Some recent ones~\cite{corona2022lisa, guo2023handnerf} use neural fields for recovering both the shape and appearance of hands. 
Yet, the heavy computation in rendering limits the usage in real-time tasks.
POEM~\cite{yang2023poem} explores multi-view mesh reconstruction on hand datasets by modeling the hand's vertices as a point cloud embedded in the intersection 
%area
volume of different camera frustums.
Nevertheless, at testing, it requires multiple synchronized views as input, which is expensive and tedious to obtain.
SiMA-Hand is a new approach, adapting multi-view hand information to boost the performance of single-view hand reconstruction, utilizing multi-view information only at training but not at testing.

\paragraph{Domain alignment.}
It has been exploited to mitigate the disparity between the source and target domains in various tasks,~\eg, image classification~\cite{yim2017gift, tung2019similarity, heo2019comprehensive} and segmentation~\cite{liu2019structured, yang2022cross}, 2D object detection~\cite{guo2021distilling, qi2021multi}, 3D object detection~\cite{wang2022sparse2dense, chen2023bevdistill},~\etc.
For RGB-based 3D hand-pose estimation, \cite{yang2019aligning, yuan2018rgb, lin2023cross} improve network performance by distilling multi-modal information such as depth maps or point clouds.
Yet, domain alignment is still under-explored in adopting multi-view knowledge to improve single-view reconstruction.
To our best knowledge, SiMA-Hand is the first attempt at single-view 3D hand reconstruction.

\section{Method}
\subsection{Overview}
\label{sec:3.1}

Fig.~\ref{fig:pipeline} shows our SiMA-Hand framework for 3D hand-mesh reconstruction.
The input to the MVR-Hand is a set of multi-view images, each of these images is taken separately as the single-view input to the SVR-Hand.
\begin{itemize}
    \item MVR-Hand effectively leverages cross-view information by fusing the image-, joint-, and vertex-level features to reconstruct a high-quality hand mesh.
    \item SiMA helps to reduce the domain disparity between the multi-view and single-view features via the hand-shape and orientation feature enhancement modules to boost the performance of the SVR-Hand. 
    % 
    % Details of the loss functions are presented in Sec.~\ref{sec:3.4}.
\end{itemize}

\paragraph{Dual-branch structure.}
Given the multi-view images of one hand, cross-view information can be better fused by disentangling the view-independent hand shape and view-dependent hand orientation.
To this end, we adopt a dual-branch structure, following~\cite{Xu_2023_CVPR}.
To be concise, we only describe the dual-branch architecture in SVR-Hand and the MVR-Hand shares the same structure. 

Specifically, given a single-view image $I$, an hourglass block is employed to extract the image-level feature $f_{i}^{S}$. 
Then, it is fed into the dual-branch network: 
(i) the shape encoder $\mathcal{F}^S_{shape}$ produces the joint-level features $f_{j}^{S}$, then a 2D-to-3D feature lifting module~\cite{chen2022mobrecon} generates the vertex-level feature $f_{v}^{S}$, followed by a SpiralConv-based~\cite{lim2018simple} decoder $\mathcal{D}^S_{shape}$ to predict the 3D vertices coordinates at the canonical pose;
and (ii) the orientation encoder $\mathcal{E}^S_{orient}$ and an MLP-based orientation regressor are employed to estimate the hand orientation.

\subsection{Multi-view Reconstruction}
\label{sec:3.2}
In the multi-view reconstruction, our MVR-Hand aims to utilize information from multiple views to obtain more accurate hand shape and orientation. 
So, we design three multi-view feature fusion modules at different levels: the image- and joint-level feature fusions for the hand orientation and the vertex-level feature fusion for the hand shape. 

\begin{figure}[t]
    \centering
    \includegraphics[width=\linewidth]{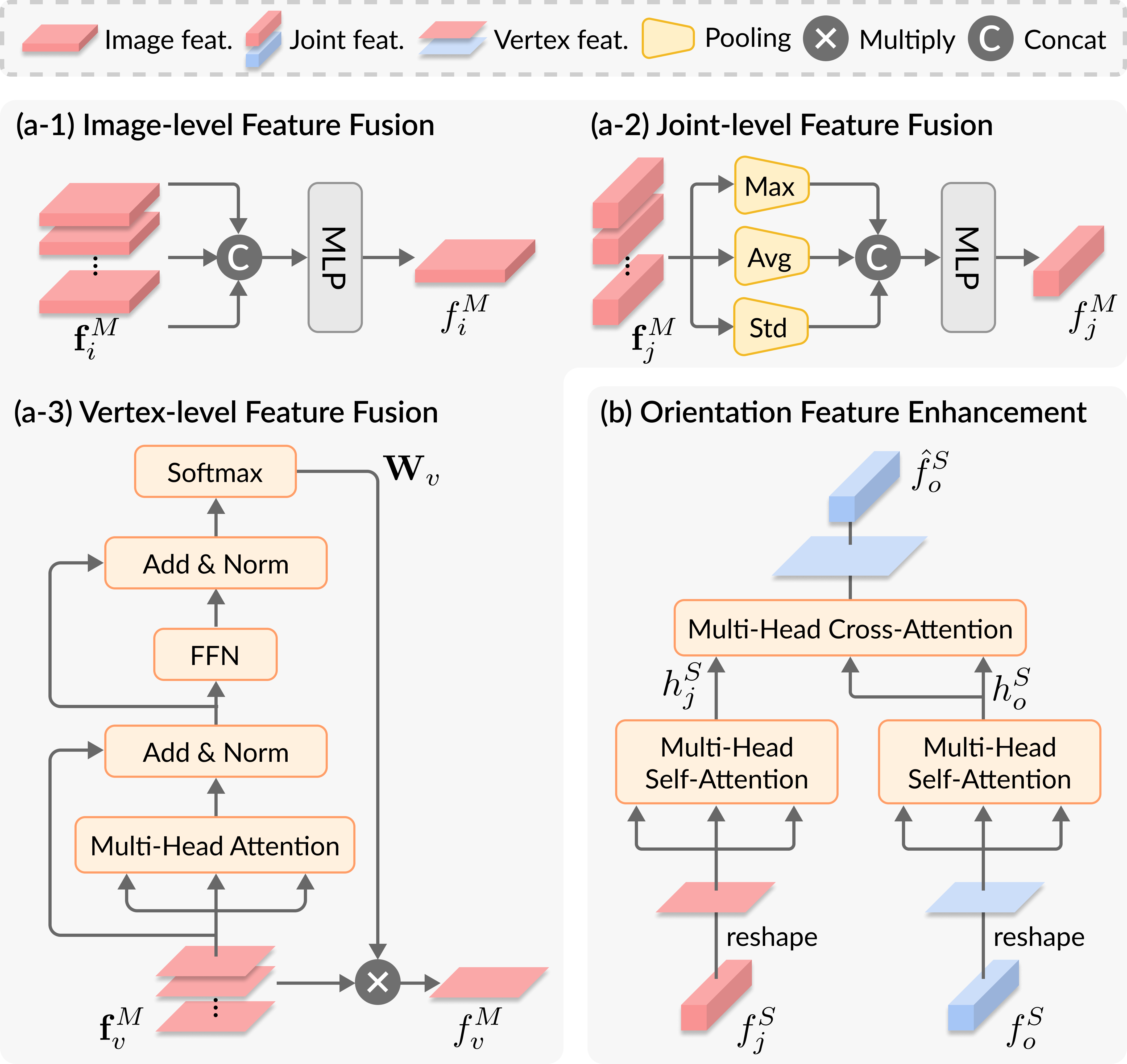}
    \caption{The designed modules in SiMA-Hand. (a) The structures of image-, joint-, and vertex-level feature fusion modules; and (b) the architecture of the orientation feature enhancement module.}
    \label{fig:modules}
\end{figure}

\paragraph{Image-level feature fusion.}
Fig.~\ref{fig:modules}(a-1) gives the structure of this module.
In detail, $\{I_k\}_{k=1}^N$ denotes $N$ perspective images of one hand.
We obtain the fused feature $f_i^M$ by feeding the concatenated image-level features $\mathbf{f}_{i}^{M}$ into the Multi-Layer Perception (MLP) $g_i(\cdot)$,~\ie,

\begin{equation}
\small
f_i^M = g_i(\operatorname{Cat}[\mathbf{f}_i^M]),
\end{equation}
where $\operatorname{Cat}[\cdot]$ denotes the concatenation operation along the feature dimension.

\paragraph{Joint-level feature fusion.} 
As a complementary, the joint-level feature $\mathbf{f}_j^M$ 
provides a global picture of the entire hand, which is a high-level guidance, for estimating the hand orientation.
To effectively fuse the information in $\mathbf{f}_j^M$, inspired by~\cite{qi2017pointnet}, we concatenate the statistics of the features pooled across different views for each joint and employ an MLP,~\ie, $g_j$, to produce the correlated joint-level feature $f_j^M$, as shown in Fig.~\ref{fig:modules}(a-2). 
Formally, we have

\begin{equation}
\small
f_j^M = g_j(\operatorname{Cat}[\operatorname{Max}[\mathbf{f}_j^M], \operatorname{Avg}[\mathbf{f}_j^M], \operatorname{Std}[\mathbf{f}_j^M]]),
\end{equation}
where $\operatorname{Max}[\cdot]$, $\operatorname{Avg}[\cdot]$, and $\operatorname{Std}[\cdot]$ denote the operations of maximum, average, and standard deviation pooling along the view dimension, respectively.

\paragraph{Vertex-level feature fusion.} 
For the hand shape, we fuse the sparse vertex-level features $\mathbf{f}_v^M$ after the 2D-3D feature lifting.
Then, we can upsample and refine the features to predict the final 3D vertex coordinates through a series of spiral convolutions. 
Considering the fact that the vertex-level features in the initial phases contain rich 3D information for recovering the hand shape at the canonical pose, the information derived from the single-view input could be heavily affected due to occlusion. 
To fuse the multi-view vertex-level features, we adopt an attention-based mechanism~\cite{vaswani2017attention} by adaptively weighing the features from different views, such that the vertices within the non-occluded regions in certain views could be emphasized in the fused feature $f_v^M$. 
The procedure is illustrated in Fig.~\ref{fig:modules}(a-3) and can be formulated as

\begin{equation}
\small
\mathbf{W}^M\!\!
= \operatorname{Softmax} (\frac{\mathbf{f}_v^M {\mathbf{f}_v^M}^T}{\sqrt{d_v}}) \mathbf{f}_v^M, \quad f_v^M = \sum_i^N{\mathbf{W}_{i}^M{\mathbf{f}_{v_i}^M}},
\end{equation}
where $\mathbf{W}_i^M$ and $\mathbf{f}_{v_i}^M$ denote the weighted score and the vertex-level feature of the $i$-th view, respectively. 
$\operatorname{Softmax}(\cdot)$ denotes the softmax operation applied on the view dimension, and $d_v$ is the feature dimension of $\mathbf{f}_v^M$.

\subsection{Single-to-Multi-view Adaptation}
\label{sec:3.3}
To avoid requiring additional views at inference, we train the single-view reconstructor,~\ie, the SVR-Hand, by distilling
% adopting 
multi-view knowledge from the MVR-Hand.

\paragraph{Hand-shape feature enhancement.} 
For the view-independent hand shape, since we directly regress the 3D vertex coordinates from $f_v^M$ at the canonical pose, directly constraining the shape feature is a simple yet effective solution. 
We formulate the hand-shape feature enhancement as the following 
% constraint 
distillation loss
between the multi-view fused feature $f_v^M$ and the single-view feature $f_v^S$,~\ie,

\begin{equation}
\small
\mathcal{L}_{SE} = ||f_v^M - f_v^S||_{2}.
\end{equation}

\paragraph{Hand-orientation feature enhancement.} 
Regarding view-dependent hand orientation, though the estimated result from multi-view features is notably precise, the 2D information utilized is highly related to the input image content, which intensely varies across different views. 
So, it is non-trivial to simulate the multi-view orientation feature directly by minimizing the distance between each pair of image- and joint-level features.
To encourage the SVR-Hand to simulate the multi-view orientation features in the latent space, we design the hand-orientation feature enhancement module that learns to enrich the orientation feature ${f_o^S}$ by the joint feature ${f_j^S}$. 
Fig.~\ref{fig:modules}(b) shows its architecture, in which we first adopt a self-attention block on flattened ${f_j^S}$ and ${f_o^S}$, separately, followed by a cross-attention block with the former as a query and the latter as the key and value,~\ie,

\begin{equation}
\small
\begin{aligned}
{h_j^S}\!\!
=\! \operatorname{Softmax}(\frac{f_j^S {f_j^S}^T}{\sqrt{d_j}})f_j^S, \;\;
{h_o^S}\!\!
=\! \operatorname{Softmax}(&\frac{f_o^S {f_o^S}^T}{\sqrt{d_o}})f_o^S, \\
{\hat{f}_o^S}
= \operatorname{Softmax}(\frac{{h_j^S} {h_o^S}^T}{\sqrt{d_o}}){h_o^S},
\end{aligned}
\end{equation}
where $d_j$ and $d_o$ are the feature dimensions of $f_j^S$ and $f_o^S$, respectively.
The enhanced feature ${\hat{f}^S_o}$ is then constrained by its counterpart ${f_o^M}$ from the MVR-Hand, \ie,
\begin{equation}
\small
\mathcal{L}_{OE} = ||f_o^M - {\hat{f}_o^S}||_{2}.
\end{equation}

\subsection{Loss Functions}
\label{sec:3.4}
Our framework is trained in two stages: pre-train the MVR-Hand, and train the SVR-Hand equipped with SiMA. 
{
\setlength{\tabcolsep}{1.5pt}
\begin{table*}[t]
    \centering
    \resizebox{\linewidth}{!}{
        \begin{tabular}{l|c|cccccc|cccccc}
        \toprule
         Methods      & Venue                       & PA-JPE$\downarrow$ & PA-JAUC$\uparrow$   & PA-VPE$\downarrow$ & PA-VAUC$\uparrow$  & PA-F@5$\uparrow$ & PA-F@15$\uparrow$ & JPE$\downarrow$ & JAUC$\uparrow$   & VPE$\downarrow$ & VAUC$\uparrow$  & F@5$\uparrow$ & F@15$\uparrow$\\
        \midrule
        Spurr~\etal & \emph{ECCV'20} & 6.83 & 86.40 & - & - & - & - & 17.34  & 69.80    & -     & -       & -    & -          \\
        METRO &  \emph{CVPR'21}             & 6.99      & -          & -        & -          & -      & -  & 15.24  & -       & -     & -       & -    & -  
        \\
        Liu~\etal & \emph{CVPR'21}        & 6.58      & -          & -        & -          & -      & -  & 15.28  & -       & -     & -       & -    & -         \\
        Tse~\etal & \emph{CVPR'22} & -      & -          & -        & -          & -      & -  & 16.05  & 72.20       & -     & -       & -    & -         \\
        HandOccNet & \emph{CVPR'22} & 5.80 & 88.40 & 5.50 & 89.00 & 77.95 & 98.99 & 14.04 & 74.82 & 13.09 & \underline{76.58} & \underline{51.49} & \underline{92.37} \\
        MobRecon & \emph{CVPR'22} & 6.36 & 87.31 & 5.59 & 88.85 & 78.48 & 98.81 & 14.20 & 73.74 & 13.05 & 76.14 & 50.83 & 92.11  \\
        H2ONet & \emph{CVPR'23} & \underline{5.65} & \underline{88.92} & \underline{5.45} & \underline{89.09} & \underline{80.14}  & \underline{99.03}  & \underline{14.02}  & \underline{74.55} & \underline{13.03}  & 76.22 & 51.29 & 92.05   \\
        SiMA-Hand & - & \textbf{5.16} & \textbf{89.37} & \textbf{4.98} & \textbf{90.04} & \textbf{81.44} & \textbf{99.24} & \textbf{13.25} & \textbf{74.99} & \textbf{12.78} & \textbf{77.12} & \textbf{53.52} & \textbf{92.48} \\
        \bottomrule
        \end{tabular}
        }
    \caption{Results on the Dex-YCB dataset. The best and second-best results are marked in bold and underlined for better comparison. Our method achieves the best performance on all metrics.}
    \label{tab:dexycb}
\end{table*}
}
First, we use the same loss function $\mathcal{L}_{Recon}$ to train the SVR-Hand and MVR-Hand for 3D hand-mesh reconstruction, which consists of several 2D and 3D loss terms.
Specifically, the 3D mesh loss $\mathcal{L}^c_M$ and 3D joint loss $\mathcal{L}^c_{J_{\scriptscriptstyle{3D}}}$ at the canonical pose are defined as

\begin{equation}
\small
\begin{aligned}
    \mathcal{L}^c_M\!=\!||\mathbf{M}^c-\mathbf{\hat M}^c||_1 \quad 
    \text{and}\ \quad \mathcal{L}^c_{J_{\scriptscriptstyle{3D}}} = ||\mathbf{J}_{c}^{\scriptscriptstyle{3D}}-\mathbf{\hat J}_{c}^{\scriptscriptstyle{3D}}||_1,\\
\end{aligned}
\end{equation}
where $\mathbf{M}$ and $\mathbf{J}^{3D}$ mean the 3D mesh and 3D joint coordinates, respectively; superscript c denotes the canonical pose; and the hat superscript denotes the ground truth. 
The same losses are computed over the predicted 3D mesh and joints after rotating by the estimated hand orientation, denoted as $\mathcal{L}^r_{M}$ and $\mathcal{L}^r_{J_{\scriptscriptstyle{3D}}}$ respectively.
The 2D joint coordinates $\mathbf{J}^{2D}$ are supervised by the normalized ground truth: 

\begin{equation}
\small
\mathcal{L}_{J_{\scriptscriptstyle{2D}}}= ||\mathbf{J}^{\scriptscriptstyle{2D}}-\mathbf{\hat J}^{\scriptscriptstyle{2D}}||_1.
\end{equation}

For the predicted hand mesh at the canonical pose, the normal and edge-length losses are employed to penalize the outlier vertices:

\begin{equation}
\small
\begin{aligned}
    &\mathcal{L}^{c}_N=\sum_{\mathbf{F} \in \mathbf{M}^{c}} \sum_{\mathbf{e} \in \mathbf{F}}|| \langle \mathbf{e},\; {\mathbf{\hat n}} \rangle ||_1,\\
   \text{and} \ \ &\mathcal{L}^{c}_E =\sum_{\mathbf{F} \in \mathbf{M}^{c}}\sum_{\mathbf{e} \in \mathbf{F}}|| |\mathbf{e}|-|{\mathbf{\hat e}}| ||_1,\\
\end{aligned}
\end{equation}
where $\mathbf{F}$ is a triangle face from the hand mesh, and $\mathbf{e}$ and $\mathbf{n}$ denote the face's edge and normal vectors, respectively. 

For the hand-orientation regression, we use the L2 loss between the output rotation matrix and the ground truth:

\begin{equation}
\small
\mathcal{L}_{R} = ||\mathbf{\hat R}{\mathbf{R}}^T - \mathds{1}||_2, 
\end{equation}
where $\mathbf{R}$ and $\mathbf{\hat R}$ are the predicted and ground-truth rotation matrix, respectively, and $\mathds{1}$ is an identity matrix.
% 

% SVR-Hand and
Overall, the loss for training the MVR-Hand is
\begin{equation}
\small
\begin{aligned}
\mathcal{L}_{Recon}
=& \mathcal{L}^c_{M}
+\mathcal{L}^c_{J_{\scriptscriptstyle{3D}}} +\mathcal{L}^r_{M}
+\mathcal{L}^r_{J_{\scriptscriptstyle{3D}}} \\
& +\mathcal{L}_{J_{\scriptscriptstyle{2D}}} 
+\mathcal{L}^{c}_N
+\mathcal{L}^{c}_E 
+\mathcal{L}_R.
\end{aligned}
\label{eq:recon_loss}
\end{equation}

Second, we train the SVR-Hand after using the single-to-multi-view adaptation with the additional supervisions based on the above-mentioned loss functions,~\ie,

\begin{equation}
\small
\mathcal{L}_{Total} = \mathcal{L}_{Recon} + \beta \mathcal{L}_{SE} + \gamma \mathcal{L}_{OE},
\label{eq:total_loss}
\end{equation}
where $\beta$ and $\gamma$ are parameters to balance the loss terms.

\section{Experiments}

\subsection{Experimental Settings}
\paragraph{Datasets.}
We conduct experiments on Dex-YCB~\cite{chao2021dexycb} and HanCo~\cite{zimmermann2021contrastive}.
Dex-YCB~\cite{chao2021dexycb} is a challenging large-scale dataset for 3D hand-object reconstruction.
It provides 1,000 sequences (over 582,000 frames) of 10 subjects grasping 20 different objects from 8 independent views.
We adopt the default ``S0'' train/test split with 406,888/78,768 samples for training/testing. 
Evaluation on this dataset can reflect the effectiveness and robustness of different methods.
HanCo~\cite{zimmermann2021contrastive} is a multi-view extension of the widely-used FreiHAND dataset~\cite{zimmermann2019freihand}. 
It contains more complicated hand gestures and more diverse objects compared to Dex-YCB.
The original training set contains 63,864 samples recorded against a green screen, whereas the testing set contains 8,576 samples from 8 different views. 
We also follow~\cite{zimmermann2019freihand} to use the official training set augmented by randomly replacing the background with 2,195 images from Flickr, yielding another 63,864 samples for training. 

\paragraph{Evaluation metrics.}
For quantitative evaluation, we adopt the commonly-used metrics following~\cite{moon2020i2l, park2022handoccnet, chen2022mobrecon,  Xu_2023_CVPR}.
JPE/VPE denotes the joint/vertex position error in millimeters (mm) calculated by the average Euclidean distance between the estimated and ground-truth 3D hand joint/vertex coordinates. 
JAUC/VAUC denotes the area under the curve of the percentage of correct keypoints (PCK) \wrt different error thresholds for joint/vertex. 
F-score is the harmonic mean of the recall and precision between the estimated and ground-truth hand-mesh vertices; we adopt F@5mm and F@15mm, which are computed with the threshold of 5mm and 15mm, respectively. 
Importantly, following existing work, we also report these metrics after Procrustes Alignment (PA), 
which directly reflects the reconstruction quality of the hand shape.

\begin{figure*}[t]
    \centering
    \includegraphics[width=\linewidth]{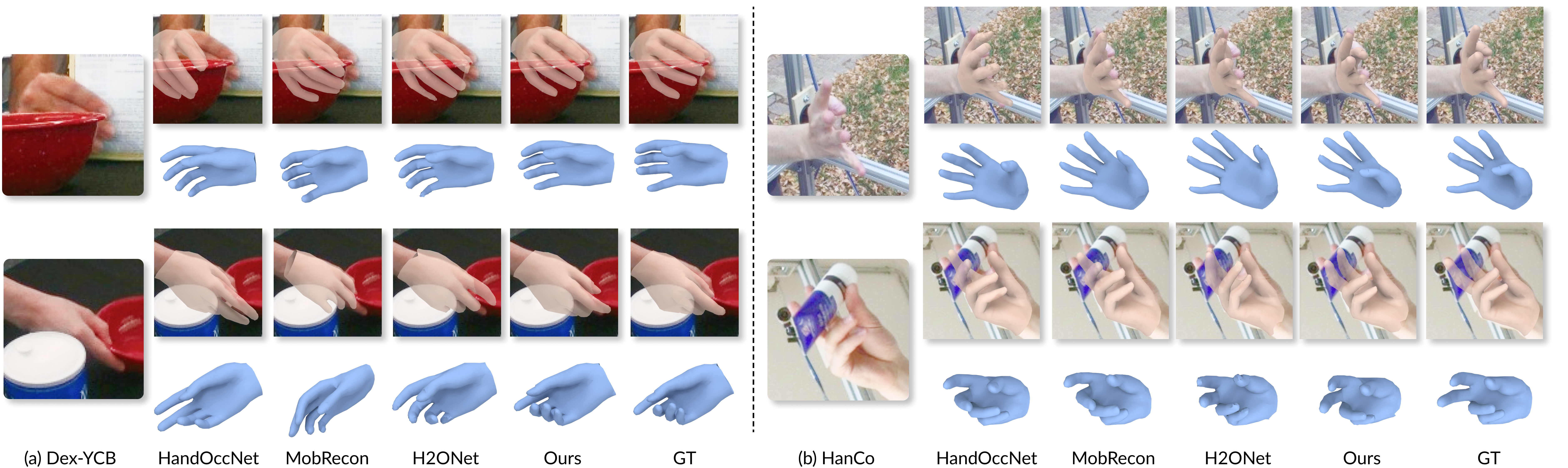}
    \caption{Qualitative comparison of our method and state-of-the-art 3D hand-mesh reconstruction methods on different datasets. The first and second rows in each example denote the normal view and another view, respectively, for better comparison.}
    \label{fig:qualitative_results}
\end{figure*}

\paragraph{Implementation details.}
We follow~\cite{chen2022mobrecon} to pre-train the feature encoder network and adopt a two-stage strategy as~\cite{Xu_2023_CVPR} to stabilize the training.
We train SiMA-Hand on four NVidia Titan V GPUs, and the Adam optimizer~\cite{kingma2014adam} is adopted.
The batch size in training is set to 64 for MVR-Hand and 128/32 for SVR-Hand.
The input image is resized to $128\!\times\!128$ and augmented by random scaling, rotating, and color jittering.
All $N\!\!=\!8$ views are used for training the MVR-Hand.
For more details, please refer to our supplementary material and code.

{
\setlength{\tabcolsep}{0.5pt}
\begin{table}[t]
    \centering
    \resizebox{\linewidth}{!}{
        \begin{tabular}{l|c|cccccc}
        \toprule
         Methods & Venue & JPE$\downarrow$ & JAUC$\uparrow$   & VPE$\downarrow$ & VAUC$\uparrow$  & F@5$\uparrow$ & F@15$\uparrow$ \\
        \midrule
        I2L-MeshNet & \emph{ECCV'20}         & 7.71 & 84.63 & 7.47 & 85.10 & 65.16 & 97.96 \\
        METRO & \emph{CVPR'21}          & 7.51 & 85.01 & \underline{6.66} & \underline{86.71} & \underline{71.78} & \underline{98.35} \\
        Liu \etal & \emph{CVPR'21}         & 8.44 & 83.19 & 8.17 & 83.71 & 62.87 & 97.07 \\
        HandOccNet & \emph{CVPR'22}         & 8.09 & 83.87 & 7.85 & 84.34 & 64.53 & 97.31 \\
        MobRecon & \emph{CVPR‘22}         & 7.95 & 84.14 & 7.12 & 85.79 & 68.45 & 98.26 \\
        H2ONet & \emph{CVPR'23}         & \underline{6.78} & \underline{86.46} & 6.89 & 86.25 & 70.61 & 98.30 \\
        SiMA-Hand  & -  & \textbf{6.55} & \textbf{86.93} & \textbf{6.48} & \textbf{87.07} & \textbf{73.16} & \textbf{98.48} \\
        \bottomrule
        \end{tabular}}
    \caption{Results on the HanCo dataset (after PA). Our method achieves the best performance on all metrics.}
    \label{tab:hanco}
\end{table}
}

{
\setlength{\tabcolsep}{1pt}
\begin{table}[t]
    \centering
    \resizebox{\linewidth}{!}{
        \begin{tabular}{r@{\hskip 2pt}|l@{\hskip 0pt}|ccc|ccc}
        \toprule
        &\multirow{2}{*}{Methods} & \multicolumn{3}{c|}{Occluded} & \multicolumn{3}{c}{Non-occluded} \\
        \cmidrule{3-8}
          & & VPE & PA-VPE  & PA-F@5 & VPE & PA-VPE & PA-F@5 \\
        \midrule
        &HandOccNet & 19.00 & 6.72 & 68.92 & 13.58 & 5.65 & 76.71 \\
        &MobRecon   & 18.48 & 6.76 & 69.47 & \underline{12.49} & 5.53 & 78.71 \\
        &H2ONet     &  \underline{18.47} & \underline{6.42} & \underline{71.73} & 12.71 & \underline{5.16} & \underline{80.76}\\
        \multirow{-4}{*}{\rotatebox{90}{Dex-YCB}} & SiMA-Hand      &  \textbf{18.29}   & \textbf{6.18}  & \textbf{73.09} & \textbf{12.40} & \textbf{4.90} & \textbf{82.01} \\
        \midrule
        &HandOccNet & 21.28 & 8.92 & 60.15 & 18.90 & 7.82 & 64.65\\
        &MobRecon   & \underline{20.08} & 8.85 & 60.68 & \underline{17.66} & 7.07 & 68.66 \\
        &H2ONet     & 20.49  & \underline{8.23} & \underline{63.72} & 18.26 & \underline{6.85} & \underline{70.80} \\
        \multirow{-4}{*}{\rotatebox{90}{HanCo}} & SiMA-Hand      &  \textbf{18.52}   & \textbf{7.67} & \textbf{67.29} & \textbf{17.27} & \textbf{6.44} & \textbf{73.32} \\
        \bottomrule
        \end{tabular}}
    \caption{Quantitative comparison between our method and recent works on occluding and non-occluding scenarios.
    }
    \label{tab:occ_vs_nonocc}
\end{table}
}

\begin{figure}[t]
    \centering
    \includegraphics[width=\linewidth]{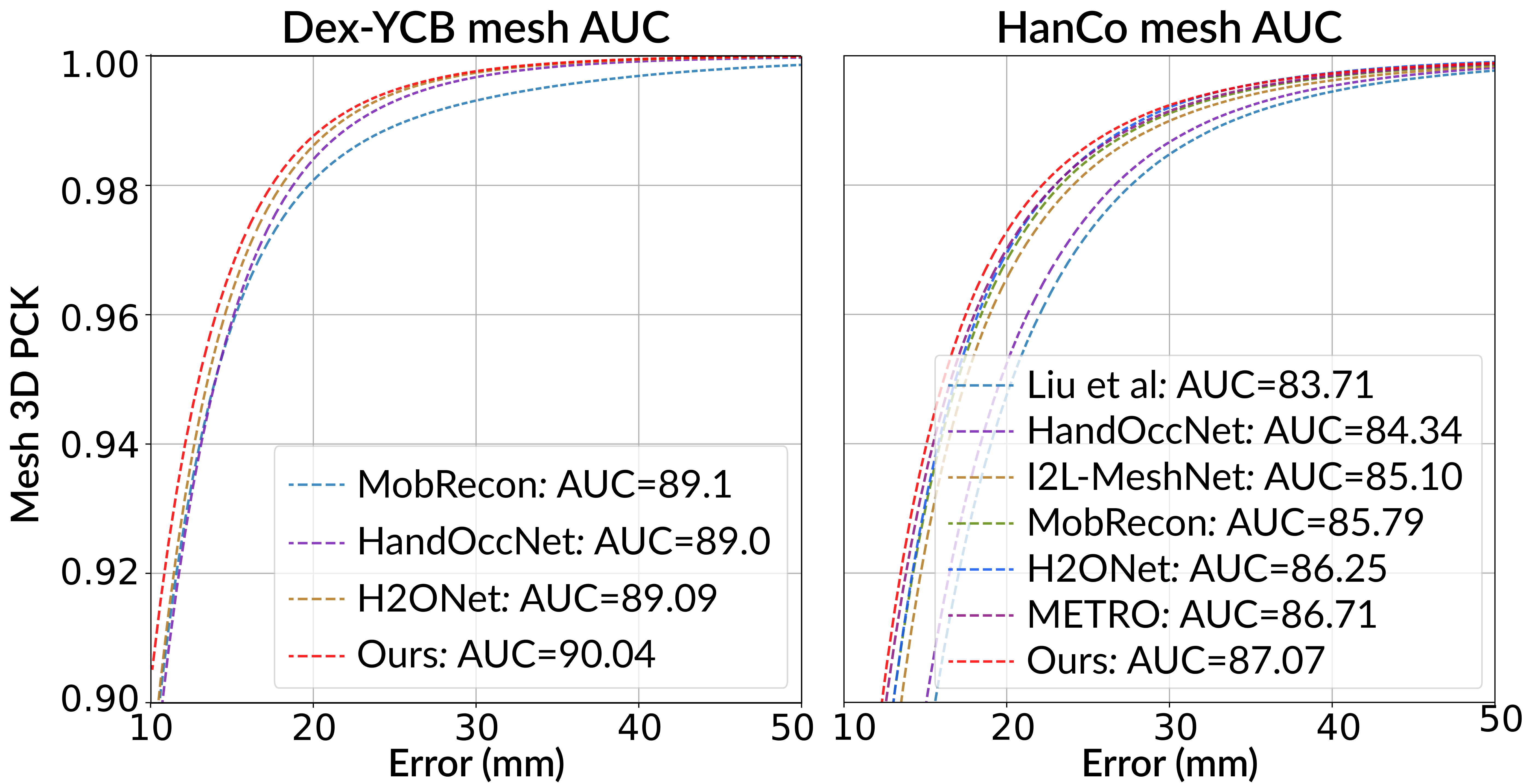}\\
    \caption{The mesh AUC comparison under different thresholds. Our method performs better than others, consistently.}
    \label{fig:pck}
\end{figure}

\subsection{Comparison with State-of-the-art Methods}
\paragraph{Evaluation on Dex-YCB.} 
We first present quantitative comparison results with the state-of-the-art methods on the Dex-YCB dataset; see Tab.~\ref{tab:dexycb} and the left plot of Fig.~\ref{fig:pck}. 
For fairness, we only report the performance of the single-frame version of H2ONet.
From the results, we can see that our SiMA-Hand achieves {\em the best results consistently} on all metrics before and after PA, showing its effectiveness both in hand-shape reconstruction and hand-orientation estimation. 
Besides, though SiMA-Hand is proposed mainly for single-view reconstruction, our MVR-Hand (PA-J/VPE=3.57/3.87) outperforms the very recent multi-view hand reconstructor~\cite{yang2023poem} (PA-J/VPE=3.93/4.00), reflecting the effectiveness of our design.
Some qualitative results are shown in Fig.~\ref{fig:qualitative_results}(a). 
Our method successfully yields plausible results, while other methods often fail in heavy occlusions.

\paragraph{Evaluation on HanCo.} 
To further evaluate the generalizability of our method, we conduct the same experiments on the HanCo dataset.
The quantitative results after PA are shown in Tab.~\ref{tab:hanco}. 
Our SiMA-Hand consistently outperforms others on all metrics. 
We also provide the mesh AUC comparison under different thresholds in the right plot of Fig.~\ref{fig:pck}. 
The visual results in Fig.~\ref{fig:qualitative_results}(b) show that our method outperforms existing works by accurately reconstructing 3D hand mesh even for challenging gestures and severe self-caused occlusions. 
More comparisons are in our supp. material.
% Please refer to our supplementary material for more quantitative and qualitative comparisons.
% 

\paragraph{Detailed evaluation on occluding and non-occluding scenarios.} 
To demonstrate the robustness of our method on challenging occluded scenes, we split the testing set of the Dex-YCB and HanCo into occluded and non-occluded parts according to the hand-level visibility as described in~\cite{Xu_2023_CVPR}. 
As shown in Tab.~\ref{tab:occ_vs_nonocc}, we obtain the best performance on both scenarios, manifesting the effectiveness of our idea.
% of single-to-multi-view adaptation. 
% 
% Please see the supplementary material for the results on all metrics.
Please see the supp. material for full metrics.

\subsection{Efficiency} 
Tab.~\ref{tab:fps} reports the inference speed (FPS), FLOPs, and the number of parameters of various methods.
The FPS is tested on an NVidia RTX 2080Ti.
SiMA-Hand is able to run in real time,
though slightly slower than MobRecon and H2ONet, it outperforms them considerably on both benchmarks. 
% 
% Also, it achieves better performance than METRO even with 25 times less computation.
% Compared to Liu~\etal and HandOccNet, although our network has more parameters due to the attention module, it still performs much fewer FLOPS with efficient graph convolution as in MobRecon.

\subsection{Ablation Studies}
We perform ablation studies on SiMA-Hand and its major components using \textbf{Dex-YCB}.
As Tab.~\ref{tab:abl_feat_fusion} shows, \textbf{IFF}, \textbf{JFF}, and \textbf{VFF} denote the image-, joint- and vertex-level feature fusions, respectively. 
\textbf{OFE} denotes the hand-orientation feature enhancement. 
\textbf{cat.} and \textbf{pool.} mean replacing feature fusion with concatenation and pooling, respectively. 
\textbf{SiMA} means adaptation between MVR- and SVR-Hand.
{
\setlength{\tabcolsep}{1.5pt}
\begin{table}[t]
\centering
    \resizebox{\linewidth}{!}{
        % \large
        \begin{tabular}{c|ccccccc}
        \toprule
        Methods & Liu \etal & HandOccNet & MobRecon & H2ONet & Ours \\ 
        \midrule
        FPS     &  32    & 30  & 66    & 52  & 48\\
        \midrule
        FLOPs (G)  &  3.66    & 15.48  & 0.46   & 0.74  & 1.65 \\
        \midrule
        \#Param. (M)  &  32.75    & 37.22  & 8.23    & 25.88  & 78.42\\
        \bottomrule
        \end{tabular}}
    \caption{Efficiency comparison with other methods.} 
    \label{tab:fps}
\end{table}
}
{
\setlength{\tabcolsep}{2.75pt}
\begin{table}[t]
    \centering
    \resizebox{\linewidth}{!}{%
    \begin{tabular}{c|c|l|ccccccccccc}
        \toprule
            & Cases &    Models     & JPE$\downarrow$ & PA-JPE$\downarrow$ & VPE$\downarrow$ & PA-VPE$\downarrow$ \\
        \midrule
        & (i) & IFF w/o & 8.92 & 3.56 & 8.88 & 3.86 \\
        % \cmidrule{3-7}
        % \cmidrule{3-7}
        & (ii) & JFF w/o & 9.48 & 3.57 & 9.43 & 3.87 \\
        & (iii) & JFF cat. & 8.55 & 3.57 & 8.54 & 3.86 \\
        % \cmidrule{3-7}
        & (iv) & VFF cat. & 9.12 & 4.08 & 8.97 & 4.21 \\
        & (v) & VFF pool. & 8.88 & 3.70 & 8.72 & \textbf{3.77} \\
        % \cmidrule{3-7}
        \multirow{-6}{*}{\rotatebox{90}{MVR}} & (vi)  & MVR-Hand & \textbf{8.49} & \textbf{3.57} & \textbf{8.48} & 3.87 \\
        \midrule
        & (vii) & OFE w/o & 13.66 & 5.23 & 13.38 & 5.36 \\
        & (viii) & OFE cat. & 13.59 & \textbf{5.22} & 13.33 & \textbf{5.35} \\
        % \cmidrule{3-7}
        \multirow{-3}{*}{\rotatebox{90}{SVR}} & (ix) & SVR-Hand & \textbf{13.37} & 5.24 & \textbf{13.11} & 5.38 \\
        \midrule
        & (x) & VFF cat.+SiMA & 13.73 & 5.39 & 13.41 & 5.49 \\
        & (xi) & VFF pool.+SiMA & 13.59 & 5.29 & 13.22 & 5.28 \\
        \multirow{-3}{*}{\rotatebox{90}{SiMA}} & (xii) & SiMA-Hand & \textbf{13.25} & \textbf{5.16} & \textbf{12.78} & \textbf{4.98} \\
         \bottomrule
    \end{tabular}%
    }
    \caption{Ablation study on major components.}
    \label{tab:abl_feat_fusion}
\end{table}
}
\paragraph{Image/joint/vertex-level feature fusion.} 
The variant-level feature fusion modules are important components in our MVR-Hand. We first analyze their impact by removing or replacing one of them.
Comparing Rows (i) and (vi), we can see that removing the IFF and directly using the target-view feature lead to degraded performance in orientation, thereby revealing the effectiveness of our IFF design.
For the JFF, comparing Rows (ii)-(iii) and (vi), either removing it or replacing the pooling-based strategy with direct concatenation will cause negative effects on JPE and VPE, demonstrating the improvement brought by our JFF. 
Similarly, we replace the attention in the VFF with pooling and concatenating-based methods to study its impact on hand-shape estimation. 
Comparing Rows (iv)-(v) and (vi), an obvious performance drop is observed, if simply using the concatenate operation to fuse features. 
Though replacing it with the pooling-based strategy obtains better PA-VPE, it is not suitable for the SiMA framework due to the inevitable information loss, as demonstrated by experiments in Rows (x)-(xii). Comparing Rows (x)-(xi) and (xii), the proposed attention-based fusion in our SiMA-Hand achieves better performance than the other two feature fusion strategies, manifesting the efficacy of our full design.

\paragraph{Hand-shape/orientation feature enhancement.}  
To assess the effectiveness of the orientation feature enhancement module, we conduct experiments in Rows (vii)-(ix) of Tab.~\ref{tab:abl_feat_fusion}. 
Comparing Rows (vii)-(viii) and Row (ix) separately, we can find that directly removing the OFF or simply concatenating the joint- and orientation feature shows subpar metrics in terms of orientation than our current design, indicating its benefit in hand-orientation estimation.
Besides, Rows (ix) and (xii) share the same network structures but without and with the feature enhancement constraints, respectively. Comparing Rows (ix) and (xii), introducing such a constraint brings a large improvement in both VPE and PA-VPE, thereby showing its necessity.

\paragraph{Distillation in SiMA.} 
We also explore different stages of distillation on the hand-shape and -orientation features in SiMA, as shown in Tab.~\ref{tab:abl_kd_loss}. 
We perform supervision on the last-layer vertex-level features in $\mathcal{D}^S_{shape}$, denoted as ${\tilde{f}}^{S}_v$. 
Comparing Rows (i) and (ii), the distillation on the vertex-level feature at the initial stage brings more improvements in hand-shape estimation than the later one, indicating its heightened potential in simulating multi-view shape features. 
Further, by fixing the hand-shape constraint on $f_v^S$, we investigate the effects of various constraints on the efficacy of hand-orientation estimation in Rows (iii)-(v). 
The result shows that distillation on $f_j^S$ has an adverse impact on the performance of hand-orientation estimation, while the one on $f_o^S$ only yields a marginal improvement.
The obvious boost in JPE and VPE comes from the distillation on the enhanced hand-orientation feature $\hat{f_o^S}$, indicating that the global information of the hand-shape features facilitates hand-orientation estimation after the fusion of $f_j^S$ and $f_o^S$.
{
\setlength{\tabcolsep}{3.25pt}
\begin{table}[t]
    \centering
    \resizebox{\linewidth}{!}{%
    \begin{tabular}{c|ccccc|cccc}
        \toprule
            & ${\tilde{f}}^{S}_v$    & $f^S_v$ & $f^S_j$ & $f^S_o$ & ${\hat{f}}^S_o$ & JPE$\downarrow$ & PA-JPE$\downarrow$ & VPE$\downarrow$ & PA-VPE$\downarrow$ \\
        \midrule
        (i) & \Checkmark & & & & & 13.90 & 5.30 & 13.39 & 5.13 \\
        (ii)  & & \Checkmark & & & & 13.79 & 5.17 & 13.29 & 4.98 \\
        % \cmidrule{2-10}
        \midrule
        (iii) & & \Checkmark & \Checkmark & & & 13.74 & 5.16 & 13.24 & 4.98 \\
        (iv) & & \Checkmark & & \Checkmark & & 13.57 & 5.18 & 13.09 & 4.99 \\
        (v) & & \Checkmark & & & \Checkmark & \textbf{13.25} & \textbf{5.16} & \textbf{12.78} & \textbf{4.98} \\
        \bottomrule
    \end{tabular}%
    }
    \caption{Ablation study on supervision in SiMA.}
    \label{tab:abl_kd_loss}
\end{table}
}

\section{Conclusion}
We introduce SiMA-Hand, a new framework for boosting the performance of 3D hand-mesh reconstruction by single-to-multi-view adaptation. To fully exploit cross-view information, we design image-, joint-, and vertex-level feature fusion. Besides, we propose feature enhancement on hand shape and orientation to leverage multi-view knowledge for tackling the occlusion issue in single-view reconstruction. Experimental results also confirm the state-of-the-art performance of SiMA-Hand on two widely-used benchmarks.

\section{Acknowledgments}
This work was supported in part by the grants from the Research Grants Council of the Hong Kong SAR, China 
(Project No. T45-401/22-N) and Project No. MHP/086/21 of the Hong Kong Innovation and Technology Fund.

\bibliography{aaai24}

\end{document}